\relax

\documentclass[letterpaper]{article}
\usepackage{aaai20}
\usepackage{times}
\usepackage{helvet}
\usepackage{courier}
\usepackage[hyphens]{url}
\usepackage{graphicx}
\usepackage{graphicx} 
\title{Explaining Neural Network Predictions for Functional Data Using Principal Component Analysis and Feature Importance}
\author{Katherine Goode, Daniel Ries, Joshua Zollweg\\ Sandia National Laboratories\\ Albuquerque, NM 87123\\ kjgoode@sandia.gov}

\usepackage{xcolor}
\definecolor{teal}{RGB}{0,153,153}
\definecolor{blue}{RGB}{0,0,255}
\definecolor{orange}{RGB}{255,140,0}

\usepackage{ulem}

\begin{document}

\maketitle

\begin{abstract}
Optical spectral-temporal signatures extracted from videos of explosions provide information for identifying characteristics of the corresponding explosive devices. Currently, the identification is done using heuristic algorithms and direct subject matter expert review. An improvement in predictive performance may be obtained by using machine learning, but this application lends itself to high consequence national security decisions, so it is not only important to provide high accuracy but clear explanations for the predictions to garner confidence in the model. While much work has been done to develop explainability methods for machine learning models, not much of the work focuses on situations with input variables of the form of functional data such optical spectral-temporal signatures. We propose a procedure for explaining machine learning models fit using functional data that accounts for the functional nature the data. Our approach makes use of functional principal component analysis (fPCA) and permutation feature importance (PFI). fPCA is used to transform the functions to create uncorrelated functional principal components (fPCs). The model is trained using the fPCs as inputs, and PFI is applied to identify the fPCs important to the model for prediction. Visualizations are used to interpret the variability explained by the fPCs that are found to be important by PFI to determine the aspects of the functions that are important for prediction. We demonstrate the technique by explaining neural networks fit to explosion optical spectral-temporal signatures for predicting characteristics of the explosive devices.
\end{abstract}

\noindent The predictive ability of machine learning models, including neural networks, makes them desirable tools in many applications including national security. However, this predictive ability comes at the cost of interpretability due to the complicated nature of the underlying algorithms. The ability to interpret a model allows users to understand how the predictions are made and assess the trustworthiness of the model. When it is not possible to directly interpret a model, it is still important to provide indirect explanations for the predictions, which especially holds true in areas with high stakes decisions such as national security.

The identification of explosive device characteristics based on optical spectral-temporal signatures obtained from videos of explosions is an example of an area in national security where machine learning could improve predictive performance. Currently, the identification of explosive device characteristics is done using heuristic algorithms or direct subject matter expert (SME) review. While machine learning may provide a more accurate identification method for this application, it is imperative that a model not only return high accuracy but an explanation for the prediction.

\begin{figure}[t]
\centering
\includegraphics[width=.95\columnwidth]{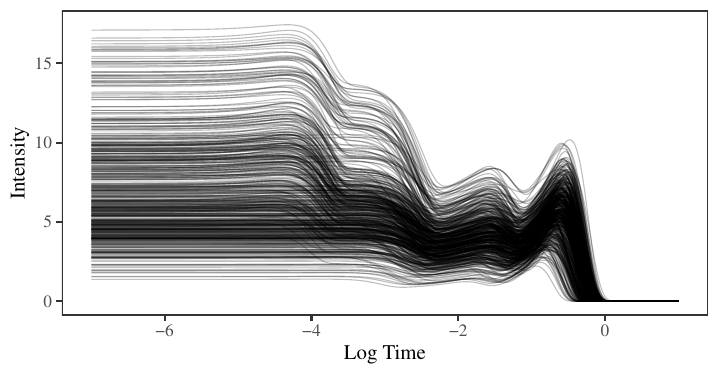}
\caption{Example of optical spectral-temporal signatures from explosions. Each function represents a signature associated with an explosive device.}
\label{fig1}
\end{figure}

The optical spectral-temporal signatures used to identify the explosive device characteristics are an example of functional data (Figure \ref{fig1}). That is, an observation corresponding to an explosive device is a function. While much research has been done relating to the explainability of machine learning models \cite{gilpin:2018,guidotti:2018,hohman:2018,molnar:2019,montavon:2018}, little work has focused on explaining machine learning models with inputs of a functional nature. 

The collection and analysis of functional data is becoming more abundant in both the government and public sectors as modern technology permits the collection of observations over dense grids in time and space (e.g., environmental sensors and wearable medical devices). A naive approach to using functional data as inputs for a machine learning model is to treat each time point as a feature in the model. Then existing explainable machine learning techniques could be applied to explain the model. However, this approach ignores the functional structure and correlation in the data, which may lead to poor predictive performance and biased explanations.

In this paper, we present a new approach for explaining predictions made by machine learning models trained using functional data that accounts for the functional aspect of the data. We combine use of the existing statistical and machine learning methods of functional principal component analysis (fPCA) and permutation feature importance (PFI) to provide insight into the characteristics of the functions that are important to the model for predictions. These identified characteristics can be presented to SMEs to determine the trustworthiness of the model and to decision makers to explain and motivate the use of the model. We demonstrate our approach by presenting an example where we explain predictions from a neural network trained using simulated optical spectral-temporal signatures from explosions to predict explosive device characteristics, which, in conjunction with an SME verification of the results, leads to trust in the model.

This paper is organized as follows. Section \ref{background} provides background on fPCA and PFI. Our procedure is introduced in Section \ref{methods} along with the motivation for the inclusion of fPCA and PFI in our method. In Section \ref{data}, we describe a data set of simulated optical spectral-temporal signatures from explosions. Section \ref{application} details the application of our approach to explain neural network predictions based on the signatures. We discuss the ability to put trust in the neural network based on the explanations produced by this approach in Section \ref{discussion} along with limitations of the approach and future research directions.

\section{Background on FPCA and PFI} \label{background}

fPCA is a common technique used in the analysis of functional data to understand the variability present in the functions \cite{ramsay:2005,wang:2016}. Similar to multivariate data principal components analysis (PCA), fPCA is a dimension reduction technique that transforms the observed data into functional principal components (fPCs). The fPCs are uncorrelated and ordered such that the first and last fPCs explain the largest and smallest amounts of variability, respectively.

PFI was originally developed by \citeauthor{breiman:2001} \shortcite{breiman:2001} for random forests, and \citeauthor{fisher:2019} \shortcite{fisher:2019} generalized the method to any predictive model. The concept of PFI is to apply a trained model to the data (training or testing) with one feature randomly permuted, and if the predictions worsen significantly when the feature is randomly permuted, the feature is considered important for prediction. In particular, a loss function is used to compare the predictions from both the permuted and non-permuted data to the true response values. A difference between the two losses is computed and summed over all observations in the data. This process is performed for all features to identify the features with the largest positive loss (i.e. the features that are most important.) Note that a negative value of PFI indicates that a randomly permuted feature results in better predictions than the observed feature. 

A disadvantage of PFI is that it is known to produce biased results when there is correlation among the features \cite{hooker:2019,nicodemus:2010,strobl:2007}. That is, simulations show that PFI reports higher importance for correlated features than is accurate. \citeauthor{hooker:2019} \shortcite{hooker:2019} hypothesize that the bias is a result the model being required to extrapolate when the permutation of a feature creates unrealistic observations in regards to the correlation between features.

\section{Methods} \label{methods}

Our original interest in using PFI as an explainability method is based on two reasons. First, we are interested in developing an approach that is capable of providing insight into the prediction process for the data analysts, the scientists, and the government decisions makers. It is likely that the familiarity with machine learning models decreases from the data analysts to decision makers. However, PFI is an easily understood explainability method that can provide information to individuals of all levels of machine learning familiarity. Data analysts can use PFI to identify features that most greatly affect prediction. The identified features can be explained to the scientists who can confirm or deny the scientific validity of the features as important to prediction. Ultimately, the aspects of the data that are identified as important to a trustworthy model can be passed on to the decision makers. The second reason we use PFI is that PFI is applicable to any type of predictive model. Hence, we could use it compare the feature importance from various machine learning models. For simplicity in this paper, we focus on neural networks.

\begin{figure}[t]
\centering
\includegraphics[width=.95\columnwidth]{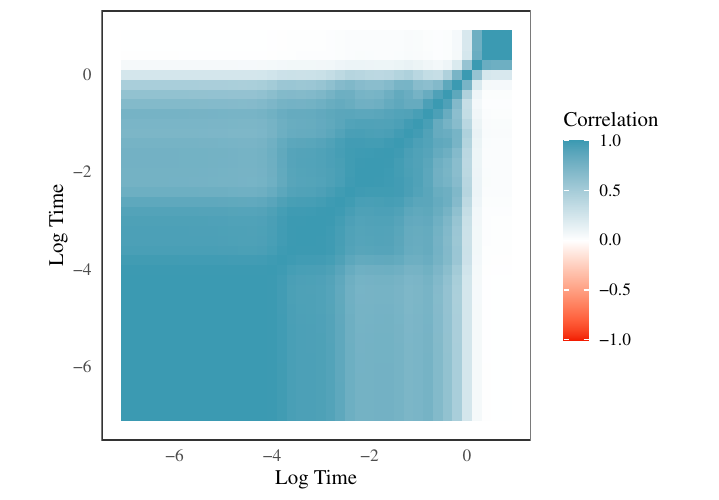}
\caption{Pearson correlations between intensity vectors at every 25th time point from simulated optical spectral-temporal explosion signatures.}
\label{fig2}
\end{figure}

As would be expected, functional data commonly exhibit large amounts of correlation between observations. Figure \ref{fig2} depicts the correlation between features in the example optical spectral-temporal explosion signatures. As a result of the correlation between features, the application of PFI to the naive approach of training a machine learning model with each time point as a feature would result in biased feature importance.

\begin{figure*}[!ht]
\centering
\includegraphics[width=1.95\columnwidth]{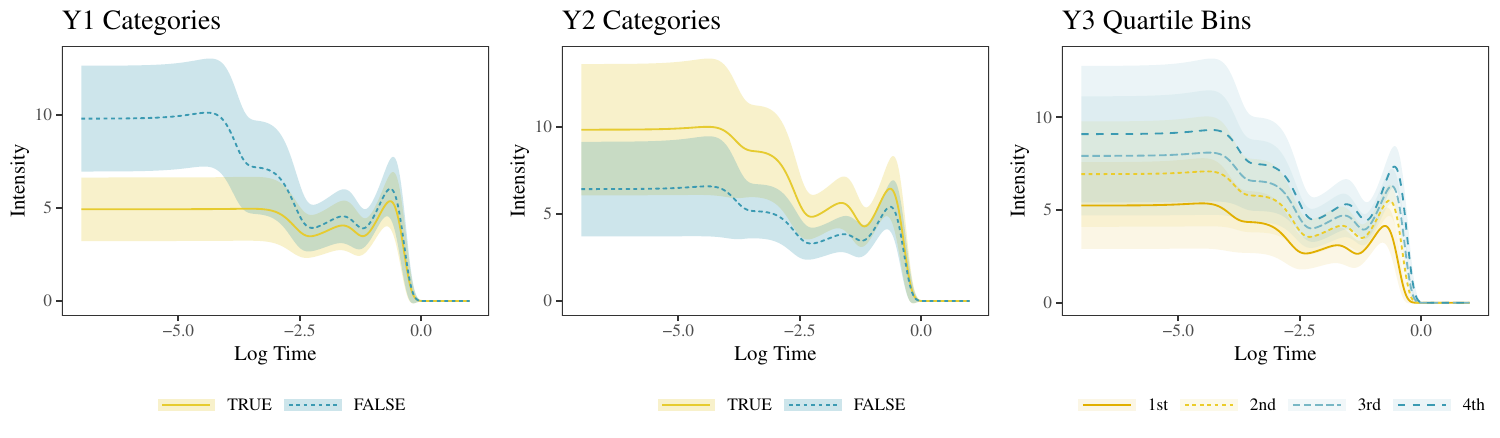}
\caption{Pointwise functional means plus/minus pointwise one standard deviations for the two categories of $Y_1$ and $Y_2$ and quartile bins of $Y_3$.}
\label{fig3}
\end{figure*}

To eliminate bias in the feature importance but still use PFI to identify the aspects of the functions important to a machine learning model for making predictions, we propose the following approach:

\begin{enumerate}
\item Remove correlation in the model features and capture the functional aspect of the data by transforming the signatures using fPCA.
\item Train a machine learning model using the uncorrelated fPCs as the features.
\item Apply PFI to identify the fPCs important for prediction without any concern of bias in the PFI due to correlation of the features.
\item Use visualizations of the fPCs to understand the variability explained by the important fPCs to identify the characteristics of the functions important for prediction.
\end{enumerate}

The visualizations of the fPCs can be presented to an SME who is able to determine if the aspects of the signatures important to the model for prediction are what is expected based on the scientific understanding of the data generating mechanism.

For the interpretation of the variability in the functions explained by the fPCs, we recommend three visualizations for considering different perspectives of the fPCs:

\begin{enumerate}
\item \textit{Eigenfunction plot}: An eigenfunction possesses weights that correspond to the times on the original scale. The magnitude of a weight indicates the importance that a time plays in the variability captured by the corresponding fPC. Larger magnitudes indicate more importance. Thus, a plot of the eigenfunction identifies the modes of variability associated with the fPC \cite{ramsay:2005}. If all weights are positive (or negative), the eigenfunction represents a weighted average of times. If there are both positive and negative weights, the eigenfunction identifies that the fPC captures a contrast between the time intervals with positive and negative values. 
\item \textit{Plot of point-wise functional mean plus/minus the eigenfunction}: Adding the eigenfunction weights times the fPC standard deviation(s) to the point-wise functional mean allows for a visualization of the principal component directions \cite{ramsay:2005}. That is, the point-wise functional mean plus/minus the eigenfunctions depicts the effect of the functional component in regards to the mean function.
\item \textit{Plot of functions with extreme fPCs}: The observed functions corresponding to the 50 highest and 50 lowest fPC values are identified and visualized along with the point-wise functional mean. The contrast in shapes of functions with high and low fPC values helps to identify the type of functional variability captured by the fPC.
\end{enumerate}

\section{Simulated Optical Spectral-Temporal Explosion Signatures} \label{data}

We consider a set of simulated signatures for the application of our explainability method. A subset of the simulated signatures is shown in Figure \ref{fig1}. A total of 10,000 signatures are created with 1,000 time points per signature. It is customary to consider the signatures on a log time scale since the events of interest occur within a short period of time (units of time intentionally excluded for anonymity).

The signatures are generated based on the scientific understanding of the relationship between three explosive device characteristics (\(Y_1\), \(Y_2\), and \(Y_3\)) and the corresponding signatures. Characteristics \(Y_1\) and \(Y_2\) are binary variables, and characteristic \(Y_3\) is a continuous variable. The characteristics affect various aspects of the signatures including the intensity, location of peaks, and number of peaks. These effects are visible in Figure \ref{fig3}, which shows the point-wise functional means for the categories of \(Y_1\) and \(Y_2\) and quartile bins for \(Y_3\) computed on the training data. In particular, \(Y_1\) affects the intensity of the signature early on, the timing of the first peak, and the total number of peaks (3 or 4). \(Y_2\) affects the intensity of the signature over the entire time. \(Y_3\) also affects the intensity of the signature over the entire time, but in addition, \(Y_3\) affects the timing of all peaks.

\begin{figure*}[t]
\centering
\includegraphics[width=1.95\columnwidth]{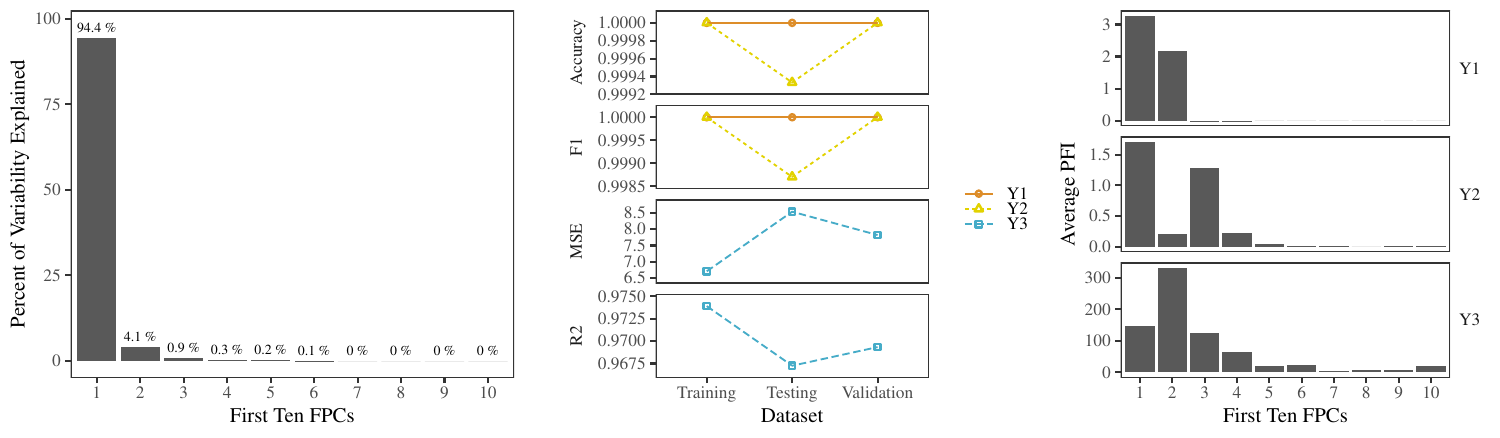}
\caption{(Left) Percent of variation explained by the first 10 fPCs. (Middle) Performance metrics for the neural networks computed on the training, testing, and validation data sets. (Right) Mean PFI values for the first 10 fPCs computed from 10 replications. The other fPCs have negligible PFI values.}
\label{fig4}
\end{figure*}

\begin{figure*}[t]
\centering
\includegraphics[width=1.95\columnwidth]{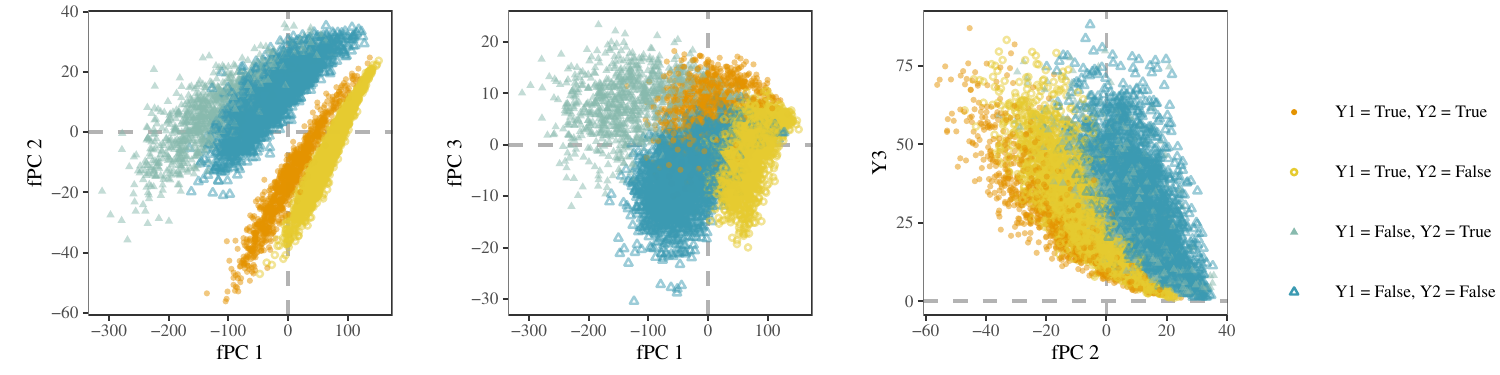}
\caption{Scatter plots depicting relationships between response variables and important fPCs identified by PFI.}
\label{fig5}
\end{figure*}

\section{Application to Explosion Optical Signatures} \label{application}

\begin{figure*}[t]
\centering
\includegraphics[width=1.95\columnwidth]{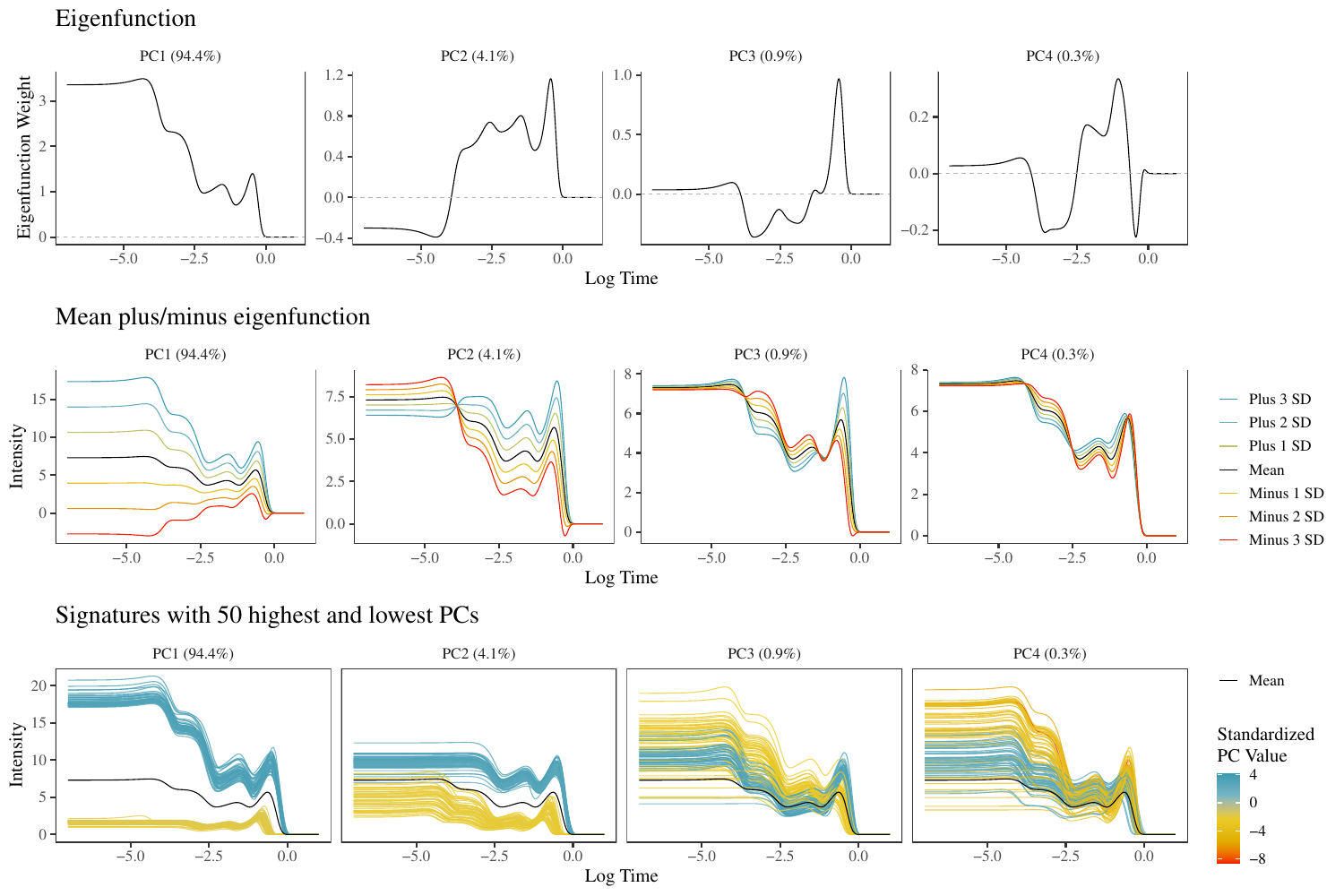}
\caption{Visualizations of the first four fPCs from the simulated optical-spectral temporal explosion signatures.}
\label{fig6}
\end{figure*}

The simulated data are randomly separated into training, testing, and validation sets containing 72.25\% (7,225), 15\% (1,500), and 12.75\% (1,275) of the signatures, respectively. Each of the 1,000 time points in the signatures are treated as a feature, and fPCA is applied to convert the 1,000 features to 1,000 fPCs. The estimated eigenfunctions from fPCA are used to transform the testing and validation data sets to fPCs. Note that in fPCA, the eigenfunctions are comparable to the eigenvectors in PCA. fPCA is performed using R (version 3.6.1; \citeauthor{r} \citeyear{r}). The percent of variability explained by the first 10 fPCs is included in Figure \ref{fig4}. The first fPC explains 94\% of the variability, and the first three fPCs combined explain 99\% of the variability. 

A neural network is trained for each of the three explosive device characteristics. It is important to keep in mind that while the majority of the variation in the data is explained by the first few principal components, this does not guarantee that the first few principal components are the best features for discriminating between explosive device characteristics. Thus, all 1,000 training data fPCs are used as features in the model. The corresponding vector of characteristics are used as the outputs. All models are fit using 3 layers with 50, 40, and 30 nodes, respectively. The transformed testing and validation features are used to assess model performance with the metrics of accuracy and F\(_1\) for \(Y_1\) and \(Y_2\) and mean squared error (MSE) and \(R^2\) for \(Y_3\). PFI is applied to the trained networks using 10 replications to account for random permutation variability. The neural networks and PFI are applied using Python (version 3.8.2; \citeauthor{vanrossum:2009} \citeyear{vanrossum:2009}) and the scikit-learn package (version 0.22.1; \citeauthor{pedregosa:2011} \citeyear{pedregosa:2011}).

All neural networks perform well (Figure \ref{fig4}). The fPCs identified as important by PFI are within the first 4 fPCs for all models (\ref{fig4}). For \(Y_1\), fPCs 1 and 2 are the most important, for \(Y_2\), fPCs 1 and 3 are the most important, and for \(Y_3\), fPC 2 is the most important with some importance for fPCs 1, 3, and 4. The fPCs greater than 10 have negligible feature importance. Since the first four fPCs are found to be the most important based on PFI, these will be the only fPCs considered for interpretation.

Figure \ref{fig5} includes scatter plots of the relationships between the explosive device characteristics and the corresponding top one or two most important fPCs identified by PFI. In all three plots, there are clear separations between \(Y_1\) and \(Y_2\) categories. For example, fPC 2 versus fPC 1 provides a clear separation between the categories of \(Y_1\) and a distinction of the categories of \(Y_2\) within the \(Y_1\) categories. The plot of \(Y_3\) versus fPC 2 shows a weak negative relationship between \(Y_3\) and fPC 2.

Visualizations for interpreting fPCs 1 to 4 are included in Figure \ref{fig6}. The visualizations of the fPCs (and all other figures) are created using the R package ggplot2 (version 3.3.0; \citeauthor{wickham:2016} \citeyear{wickham:2016}). The interpretation of fPCs 3 and 4 are not as clear as fPCs 1 and 2 since fPCs 3 and 4 explain a smaller amount of variability. However, the variability explained by all fPCs that PFI identifies as important for prediction can be connected to the effects caused by the explosive device characteristics. The fPCs are interpreted as follows:

\begin{itemize}
\item \textit{fPC 1}: The eigenfunction for fPC 1 makes it clear that fPC 1 is a weighted average over all times. The visualizations of the point-wise mean function plus/minus the eigenfunctions and the signatures corresponding to the 50 highest and lowest fPC 1 values indicate that fPC 1 captures a contrast between signatures with high intensity starting values, a large decrease in intensity over time, and four peaks and signatures with low starting values, a relatively constant value over all times, and three peaks. 
\item \textit{fPC 2}: The eigenfunction for fPC 2 makes it clear that fPC 2 explains a contrast between time points before and after -3.75. The visualizations of the point-wise mean function plus/minus the eigenfunctions and the signatures corresponding to the 50 highest and lowest fPC 2 values indicate that fPC 2 captures a contrast between signatures with high starting values and 3 peaks that occur after the mean function and signatures with lower starting values and four peaks before the mean function. 
\item \textit{fPC 3}: The eigenfunction of fPC 3 indicates that the fPC explains a contrast in variability between the times of (-3.75, -1.75) and (-1.75, 0). The mean plus/minus eigenfunctions and signatures with extreme fPC 3 values suggest that fPC 3 captures the variability between signatures with lower values between the first time interval and a large fourth peak during the second time interval and signatures with higher values during the first time interval and a small fourth peak during the second time interval. 
\item \textit{fPC 4}: The eigenfunction for fPC 4 depicts that fPC 4 explains a contrast between the two time intervals of (-4,-2.5) and (-0.5,0) and the time interval of (-2.5,-0.5). The other two visualizations for fPC 4 indicate that fPC 4 captures the variability between signatures with a steep decrease during the time interval of (-4,-2.5) and dramatic third and fourth peaks during the time intervals of (-2.5,-0.5) and (-0.5,0), respectively and signatures with less intense peaks throughout the entire time interval of (-4,0). 
\end{itemize}

To connect the interpretations of the first four fPCs to the explosive device characteristics, consider the functional means for the \(Y_1\), \(Y_2\), and \(Y_3\) in Figure \ref{fig3}. PFI identified fPCs 1 and 2 as being important for predicting \(Y_1\). This is reasonable since both fPCs capture a variability in functions with intensity during early times, different timings for peak 1, and the number of peaks (3 or 4). PFI found that fPCs 1 and 3 are important for predicting \(Y_2\), which is reasonable since both fPCs capture a variability between signatures with high and low intensities across the entire time interval. PFI identified fPC 2 as being the most important for predicting \(Y_3\), which is reasonable since fPC 2 captures the variability between signatures with high intensity values and peaks occurring after the mean function and signatures with low intensity values and peaks occurring before the mean function. PFI also identified fPCs 1, 3, and 4 as having some importance for predicting \(Y_3\), and these fPCs pick up on smaller amounts of variability affected by \(Y_3\) such as the intensity in certain regions and the intensity of the fourth peak. By sharing these findings with an SME, we are able to confirm that the fPCs identified by PFI as important capture the type of variability in the signatures that is important to the corresponding explosive device characteristics.

\section{Discussion} \label{discussion}

\begin{figure}[t]
\centering
\includegraphics[width=.95\columnwidth]{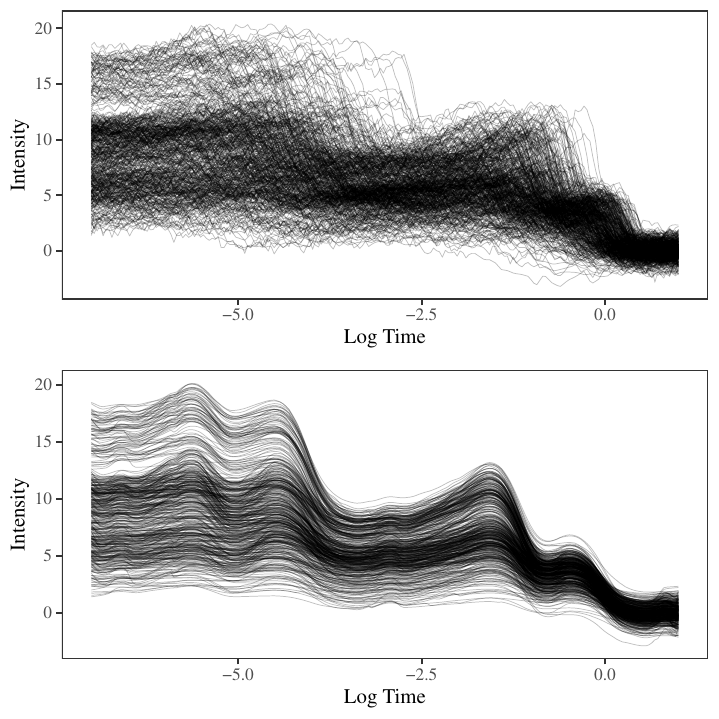}
\caption{(Top) Examples of simulated optical spectral-temporal signatures from explosions with more variability. (Bottom) The signatures from the plot on the left after applying smoothing and alignment (using box filtering and time warping, respectively) from the fdasrvf R package (version 1.9.3; \citeauthor{tucker:2020} \citeyear{tucker:2020}).}
\label{fig7}
\end{figure}

The prediction of explosive device characteristics using optical spectral-temporal signatures from explosions is an example in national security where machine learning applied to functional data could improve performance in practice. However, this is also an example of a machine learning application where not only is high predictive accuracy important, but it is also imperative that it is possible to explain how the model makes predictions. In this paper, we propose a procedure for training and explaining machine learning models with functional data inputs that accounts for the functional nature of the data. We implement our procedure to provide explainable predictions for neural networks trained to predict explosive device characteristics. In particular, the transformation of the optical spectral-temporal signatures using fPCA permits the identification of fPCs important to prediction in a neural network for an explosive device characteristic using PFI, and visualizations for interpreting the variability captured by the important fPCs allows for the determination of the aspects of the signatures that are important for prediction. The validation from an SME of the meaningfulness of the fPCs identified by PFI allows us to be confident that the neural networks are using trustworthy aspects of the signatures to make predictions.

A limitation of this method is that the ability to explain a prediction made by the neural network is dependent on the ability to interpret the fPCs. In our example, PFI identifies the first four fPCs as important for predicting at least one of the characteristics, and it is possible to determine meaningful variation captured by these fPCs. However, if PFI identifies fPCs that are not able to be interpreted, it would not be possible to explain the aspects of a function that are important to the neural network for prediction. While the earlier fPCs explain larger amounts of the variability in a data set, it is not necessarily true that the earlier fPCs will be the best for discrimination of response characteristics. If PFI identifies a higher numbered fPC as important, it is likely to be more difficult to interpret.

Another aspect not considered in this paper is that fPCA accounts for amplitude variability (vertical variability) but does not account for phase variability (horizontal variability) in the functions. Joint functional principal component analysis (joint fPCA) is a method that can be applied after smoothing and aligning functional data that accounts for both amplitude and phase variability \cite{lee:2017,tucker:2013}. The procedure in this paper could be adjusted by substituting fPCA with smoothing, aligning, and applying joint fPCA to the signatures (Figure \ref{fig7}). With noisier signatures, accounting for phase variability is important to capture the signals in the data.

The focus of this paper is the explanation of the predictions. However, another aspect important to applications connected to high consequence decisions is the ability to report uncertainty quantification to gauge the model's confidence in a prediction. Bayesian neural networks (BNNs) are an example of a machine learning method that returns uncertainty quantification. BNNs produce a distribution for prediction as opposed to a single prediction. If a BNN is used as the machine learning model in our procedure, it would be possible to develop a method that adjusts the computation of PFI to account for the distribution of predictions.

In this paper, we present a method for explaining predictions from a machine learning model trained using functional data. We demonstrate the method on a national security application in which model authentication is crucial. While all statistical methods used within our procedure have been developed previously, we join these techniques in a new way that provides insight to a black-box model. Additionally, our approach accounts for the functional nature of the data, which is an aspect that has been overlooked in the explainable machine learning literature.\\
\\
\textit{Sandia National Laboratories is a multimission laboratory managed and operated by National Technology \& Engineering Solutions of Sandia, LLC, a wholly owned subsidiary of Honeywell International Inc., for the U.S. Department of Energy's National Nuclear Security Administration under contract DE-NA0003525. This paper describes objective technical results and analysis. Any subjective views or opinions that might be expressed in the paper do not necessarily represent the views of the U.S. Department of Energy or the United States Government. SAND2020-11247 C}

\section*{Acknowledgments}

The authors would like to thank J. Derek Tucker for his advice on fPCA.

\bibliographystyle{aaai}
\bibliography{bibliography.bib}

\end{document}